\pdfoutput=1

\documentclass[11pt]{article}

\usepackage[]{acl}

\usepackage{times}
\usepackage{latexsym}
\newcommand{\modelname}{DIONYSUS}
\newcommand{\modelnamebase}{DIONYSUS\textsubscript{BASE}}
\newcommand{\modelnamelarge}{DIONYSUS\textsubscript{LARGE}}
\definecolor{mygray}{gray}{.9}
\usepackage{multirow}
\usepackage{mathtools}
\usepackage{tablefootnote}
\usepackage{algorithm}
\usepackage{algpseudocode}
\usepackage{amsmath}
\usepackage{xcolor}
\usepackage{colortbl}

\DeclareMathOperator*{\argmax}{\arg\!\max}

\usepackage{booktabs}
\usepackage{graphicx}
\usepackage[T1]{fontenc}

\usepackage[utf8]{inputenc}

\usepackage{microtype}

\title{\modelname: A Pre-trained Model for Low-Resource Dialogue Summarization}

\author{Yu Li\thanks{~~Work was done when Yu Li was interning at MSR}~$^\dag$, Baolin Peng$^\ddag$, Pengcheng He$^\ddag$, Michel Galley$^\ddag$, Zhou Yu$^\dag$, Jianfeng Gao$^\ddag$ \\
  $^\dag$Columbia University, New York, NY \\
  $^\ddag$Microsoft Research, Redmond, WA \\
  \texttt{\{yl5016, zy2461\}@columbia.edu} \\
  \texttt{\{bapeng,penhe,mgalley,jfgao\}@microsoft.com} 
}

\begin{document}
\maketitle
\begin{abstract}
Dialogue summarization has recently garnered significant attention due to its wide range of applications. However, existing methods for summarizing dialogues have limitations because they do not take into account the inherent structure of dialogue and rely heavily on labeled data, which can lead to poor performance in new domains. In this work, we propose \modelname\ (dynamic input optimization in pre-training for dialogue summarization), a pre-trained encoder-decoder model for summarizing dialogues in any new domain. To pre-train \modelname, we create two pseudo summaries for each dialogue example: one from a fine-tuned summarization model and the other from important dialogue turns. We then choose one of these pseudo summaries based on information distribution differences in different types of dialogues. This selected pseudo summary serves as the objective for pre-training \modelname\ using a self-supervised approach on a large dialogue corpus. Our experiments show that \modelname\ outperforms existing methods on six datasets, as demonstrated by its ROUGE scores in zero-shot and few-shot settings.
\end{abstract}

\section{Introduction}
\begin{figure}[htb]
    \centering
    \includegraphics[trim={0 0 0 0},clip,width=7.5cm]{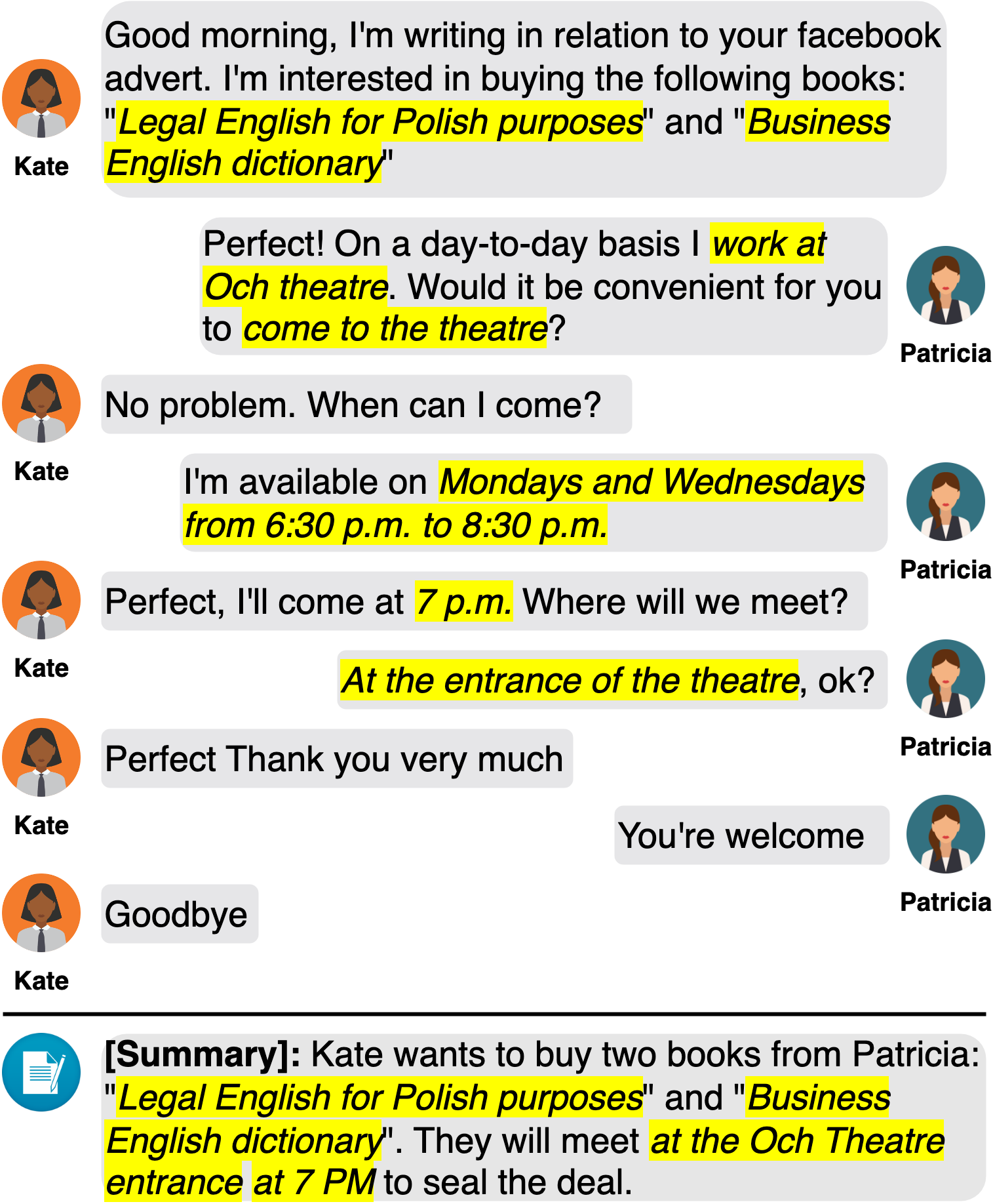}
    \caption{A summary of a dialogue in the SAMSum dataset, where the golden summary effectively compiles relevant information (in yellow) from the entire conversation.}
    \label{fig:intro}
    \vspace{-3mm}
\end{figure}
Text summarization aims to produce concise and accurate summaries of long texts. Recent research on pre-trained neural language models has shown success in summarizing monologues \cite{lewis-etal-2020-bart, Raffel-t5-2022, zhang2019pegasus, he2022z}, such as news articles \cite{lee-etal-2022-neus, ravaut-etal-2022-summareranker} and scientific publications \cite{IBRAHIMALTMAMI20221011, dong-etal-2021-discourse}. However, dialogue summarization presents additional challenges due to the different information distribution in dialogues.

Self-supervised text summarization models \cite{zhang2019pegasus, wan-bansal-2022-factpegasus, Phang2022InvestigatingEE} are typically pre-trained on free-form text data, with selected sentences as the pre-training objective. While this approach can be effective for monologues such as news articles, it is less successful at summarizing semistructured and multi-participant dialogues. As illustrated in Figure \ref{fig:intro}, in daily chats, dialogue information is often dispersed across various dialogue turns, making it difficult to extract all relevant information through a few selected turns. While a golden summary needs to accurately captures vital information throughout the entire conversation. Furthermore, real-world dialogue-summarization applications often have limited or even no labeled data, making it challenging to develop effective models. Therefore, it is crucial to develop dialogue summarization models that can perform well in zero-shot and few-shot settings for their practical use.

To address these challenges, we propose \modelname, a pre-trained sequence-to-sequence model designed to summarize dialogues in any domain, even with a lack of labeled data. It uses pseudo summaries as its pre-training objective, which can be dynamically selected from two sources.

First, for daily chats where multiple dialogue turns are not sufficient to summarize the dialogue, we train a summary helper using high-quality dialogue summarization datasets to generate pseudo summaries for these types of dialogues. On the other hand, for dialogues like meeting minutes, interviews, and debates, which can be summarized through a selection of essential turns, we use a method inspired by the gap sentence generation (GSG) technique in PEGASUS to select these turns as pseudo summaries for training. For instance, choosing the final few turns in a conversation can effectively summarize meeting minutes. We have improved upon the GSG method by using the generated summaries from the summary helper as references during gap sentence selection, as they tend to have less noise compared to the full dialogue context. We refer to this source of pseudo summaries as ``Principal'' and refer to our improved method as GSG+. We find that our improved method outperforms previous methods in low-resource settings across different domains, such as daily chats, emails, and customer service dialogues. Additionally, we study different objective strategies for selecting the pseudo summary as a pre-training objective from the generated summary and the ``Principal.''

We evaluate \modelname\ on six dialogue summarization datasets. Our best model trained on 19 dialogue corpora surpasses PEGASUS\textsubscript{LARGE} in a zero-shot setting across all domains. We also found that the best performance is achieved by selecting the source with the highest ROUGE score as the objective strategy. Our main contributions are:
\begin{itemize}
    \item The development of \modelname, a pre-trained sequence-to-sequence model for summarizing dialogues in any domain in a zero-shot or few-shot setting.
\end{itemize}
\begin{itemize}
    \item The introduction of new self-supervised pre-training objectives for dialogue summarization using a summary helper and GSG+.
\end{itemize}
\begin{itemize}
    \item The demonstration that \modelname\ outperforms baselines on six domains in low-resource settings, and can be fine-tuned with only 10 training examples to outperform vanilla T5 \cite{Raffel-t5-2022} fine-tuning with $1,000$ examples.
\end{itemize}

\section{Approach}
\begin{figure*}[ht]
    \centering
    \includegraphics[trim={0 0 0 0},clip,width=16cm]{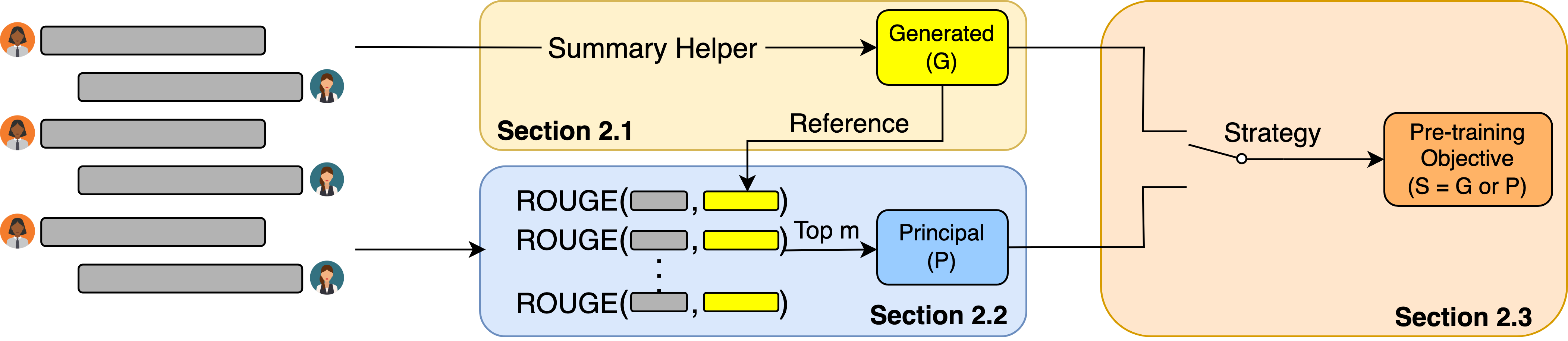}
    \caption{A diagram of pre-training in \modelname: The summary helper ($\S$ \ref{sec:helper}) generates a pseudo-summary (G) to select dialogue turns ($\S$ \ref{sec:principals}) as the ``Principal'' (P) and using various strategies ($\S$ \ref{sec:strategy}) to choose between the generated summary and the principal as the pre-training objective.}
    \label{fig:method}
    \vspace{-3mm}
\end{figure*}
Figure \ref{fig:method} outlines the steps for constructing \modelname: $\S$ \ref{sec:helper} First, a summary helper is constructed using two high-quality dialogue summarization datasets. This helper generates a pseudo summary for each dialogue in our pre-training corpus. $\S$ \ref{sec:principals} Next, the ``Principal'' is extracted using GSG+ as the other pseudo summary for the dialogue. $\S$ \ref{sec:strategy} Finally, various strategies are employed to select the best pseudo summaries from the first and second steps to serve as the objective for pre-training.

\subsection{Summary Helper}
\label{sec:helper}
In certain types of dialogue, such as daily chats, it can be challenging to gather all necessary information from just a few dialogue turns due to the dispersed nature of dialogue information. To address this problem, we have created a summary helper model that generates pseudo summaries for each training example in our pre-training corpus.

We build our summary helper upon the T5 \cite{Raffel-t5-2022} model. To capture essential information in a dialogue, we have trained our helper on the MultiWoz dataset \cite{budzianowski2018large, eric-etal-2020-multiwoz} in DS2 \cite{shin-etal-2022-dialogue}, which contains summaries derived from dialogue states using templates. This allows us to capture essential information from each turn in the conversation. Additionally, we have continued training our helper on the DialogSum \cite{chen-etal-2021-dialogsum} dataset, a human-annotated dataset in the daily life domain. This allows us to overcome the fixed format of summaries introduced by templates in DS2 and produce more natural pseudo summaries.

\subsection{Gap Sentence Generation Plus (GSG+)}
\begin{algorithm}
\caption{GSG+}
\label{algo:principals}
\begin{algorithmic}[1]
\State $P \gets \emptyset$
\For{$j \gets \text{1 to m}$}
    \State $s_i := rouge(P \cup \{x_i\}, G), \forall i\;s.t.\;x_i \notin P$
    \State $k := \argmax \{s_i\}_n$
    \State $P := P \cup \{x_k\}$
\EndFor
\end{algorithmic}
\end{algorithm}
\label{sec:principals}
Dialogues in certain settings, such as meetings and medical dialogues, often include summary turns that summarize the entire conversation. For example, a participant may summarize a meeting, or a doctor may explain the outcome. These summary turns can be used as a pre-training objective because they highlight the main points of the dialogue and provide a concise overview of the topic discussed. In order to make \modelname\ more adaptable to these scenarios, we have improved the independent principal method in the GSG method \cite{zhang2019pegasus} by using it to select essential summary turns as pseudo summaries for training. Our new method, called Gap Sentence Selection Plus (GSG+), uses the ROUGE1-F1 score between each dialogue turn $x_i$ and the generated summary $G$ from the helper in Section \ref{sec:helper} rather than the remaining text $D \setminus x_i$ to determine the importance of each turn. The generated summary eliminates much of the extraneous information from the dialogue and thus tends to have less noise than the full dialogue context, resulting in a less cluttered summary. This enables us to select the top-m-scored summary turns as the ``Principal,'' which will provide a more comprehensive overview of the vital information in the dialogue. For instance, Using the summary helper to identify key points increases the likelihood of selecting the most important dialogue turns as the ``Principal'' summary when creating pseudo summaries for meeting minutes instead of randomly selecting dialogue turns.

Specifically, given a dialogue $D = \{x_i\}_n$, we use Algorithm \ref{algo:principals} to obtain the pseudo-summary ``Principal'' $P$. The input for our training example is the remainder of the dialogue $D \setminus P$. In Appendix \ref{sec:effect of order}, we explore the impact of the dialogue turns order on the formation of the ``Principal''. Using GSG+ can effectively identify essential summary turns and generate more accurate pseudo-summaries than with the original GSG method.

\subsection{Pre-training Objectives Strategy}
\label{sec:strategy}
\begin{algorithm}
\caption{Better ROUGE}
\label{algo:better_rouge}
\begin{algorithmic}[1]
\State $S \gets \emptyset$
\State $s_g := rouge(G, D \setminus \{P\})$
\State $s_p := rouge(P, D \setminus \{P\})$
\If{$s_g > s_p$}
    \State $S := G$
\Else
    \State $S := P$
\EndIf
\end{algorithmic}
\end{algorithm}
To generate the final pseudo summary $S$ for each specific dialogue training example, we consider three strategies. These strategies are based on the generated pseudo summary $G$ and the extracted ``Principal'' $P$. These strategies serve as the pre-train objective for the dialogue training example.

\paragraph{All G}
$S = G$: We always select the generated summary from the summary helper as the pre-training objective.

\paragraph{All P}
$S = P$: We always select the ``Principal'' as the pre-training objective.

\paragraph{Better ROUGE}
We use either $G$ or $P$ based on the recall of information from the dialogue to determine the pre-training objective. We utilize Algorithm \ref{algo:better_rouge} to get the pre-training objective by calculating the ROUGE1-F1 score for the pseudo summaries and the dialogue, excluding the ``Principal'' $D \setminus P$. It is important to note that we use the same reference to ensure a fair comparison.

For pre-training with above strategies, if we choose $G$ as the pseudo summary, we input the full dialogue. If we choose $P$, we input the dialogue, excluding the ``Principal,'' $D \setminus P$ to create an abstract summary. However, we also include the ``Principal'' with a probability, using a copying mechanism to create an extractive summary. More information about this copy mechanism can be found in Section \ref{sec:copying mechanism}. It is important to note that we do not combine these two pseudo summaries for a single training example. Each example in our pre-training corpus will have either $G$ or $P$ as its designated pseudo summary.

\section{Training Corpus}
To train \modelname, we utilized 19 conversational corpora that do not come with pre-defined dialogue summaries. We employed a self-supervised approach by using pseudo-summaries as the pre-training objective.

\paragraph{Conversational Corpora} We collect 19 available conversational corpora consisting of 1.7M examples after truncating for pre-training. Corpus information is listed in Table \ref{tab:pre training corpora}. We access these corpora through ConvoKit v2.5.3\footnote{https://convokit.cornell.edu/}. This helps us to ensure that \modelname\ is well-equipped to handle a variety of conversational scenarios.

\begin{table}[h!]
\centering
\small
\begin{tabular}{lr}
    \toprule
    \textbf{Corpora}  & \textbf{\# Dialogues}\\
    \midrule
     CaSiNo \cite{chawla-etal-2021-casino} & 1,030\\
     Chromium \cite{meyers-etal-2018-dataset} & 163,675\\
    Gone Awry (CMV) \cite{zhang-etal-2018-conversations} & 6,842\\
    Gone Awry (Wiki) \cite{zhang-etal-2018-conversations} & 4,188\\
    Diplomacy \cite{peskov-etal-2020-takes} & 246\\
    Friends \cite{zhou-choi-2018-exist} & 1,301\\
    GAP \cite{braley-2018-gap-corpus} & 28\\
    IQ2 \cite{zhang-etal-2016-conversational} & 108\\
    Cornell Movie Dialogs\tablefootnote{Cornell Movie Dialogs Corpus is from \citet{danescu-niculescu-mizil-lee-2011-chameleons}} & 83,097\\
    Parliament \cite{zhang-etal-2017-asking} & 216,894\\
    \textsc{PersuasionForGood}\tablefootnote{\textsc{PersuasionForGood} is from \citet{wang-etal-2019-persuasion}} & 1,017\\
    Reddit Coarse \cite{Zhang_Culbertson_Paritosh_2017} & 9,483\\
    Reddit Corpus (small) \tablefootnote{https://convokit.cornell.edu/documentation/reddit-small.html} & 8,286\\
    Supreme Court \tablefootnote{https://convokit.cornell.edu/documentation/supreme.html} & 7,700\\
    Switchboard \cite{stolcke-etal-2000-dialogue} & 1,155\\
    Tennis \cite{fu2016tie} & 81,974\\
    Wiki Deletion \cite{Mayfield-2019-analyzing} & 383,918\\
    Wiki Talk Pages\tablefootnote{Wikipedia Talk Pages is from \citet{Danescu-Niculescu-Mizil+al:12a}} & 125,292\\
    Winning Arguments \cite{tan-2016-winning} & 3,051\\

    \bottomrule
\end{tabular}
\caption{Corpora we use to pre-train \modelname.}
\label{tab:pre training corpora}
\vspace{-3mm}
\end{table}

We train our objective summary helper with a rule-based dialogue summarization dataset (DS2) and an abstractive summarization dataset (DialogSum).

\paragraph{DS2} This dataset \cite{shin-etal-2022-dialogue} creates dialogue summaries for the MultiWOZ \cite{budzianowski2018large, eric-etal-2020-multiwoz} dataset by heuristic rules from the dialogue states. It includes 5 domains and $10,000$ dialogues.

\paragraph{DialogSum} This dataset \cite{chen-etal-2021-dialogsum} collects human annotated summaries for daily-life dialogues from three datasets: DailyDialog \cite{li-etal-2017-dailydialog}, DREAM \cite{sundream2018}, and MuTual \cite{cui-etal-2020-mutual}, as well as dialogues from an English-speaking practice website. It has 13,460 dialogues in total.

\section{Experiments}
\subsection{Downstream Tasks and Metrics}
We evaluate our methods on three public dialogue summarization datasets or benchmarks: SAMSum \cite{gliwa-etal-2019-samsum}, ConvoSumm \cite{fabbri-etal-2021-convosumm}, and \textsc{TweetSumm} \cite{feigenblat-etal-2021-tweetsumm-dialog}

\paragraph{SAMSum} This dataset contains over 16k natural messenger-like dialogues with manually annotated summaries by language experts. 

\paragraph{ConvoSumm} It is a benchmark of four domains: New York Times comment, StackExchange, W3C email, and Reddit. Dialogues are extracted from publicly available data, and each domain has 500 dialogues. They hire crowdsorce workers on Amazon Mechanical Turk to annotate dialogue summary.

\paragraph{TweetSumm} This dataset contains 1,100 reconstructed real-world customer support dialogues from Tweet. Each dialogue has human annotated abstractive summaries and extractive summaries. We only use abstractive summaries in the dataset as references in our experiments.

We report ROUGE-1, ROUGE-2, and ROUGE-L scores \cite{lin-2004-rouge} to evaluate generated summaries against references.

\subsection{Baselines}
We compare our methods with three competitive baselines.

\paragraph{T5v1.1} It is an improved version of the original T5 model \cite{Raffel-t5-2022}. Since the original T5 model is pre-trained on downstream tasks in supervised learning, the test set of downstream tasks overlaps with the pre-training data. To make a fair comparison in a zero-shot setting, we choose T5v1.1 as it is pre-trained on C4 without mixing in the downstream tasks.

\paragraph{PEGASUS} \citet{zhang2019pegasus} propose this pre-trained model for abstractive summarization tasks. The pre-training objective is GSG, transforms any text into an abstractive summarization example by selecting important sentences as output summaries. We use the PEGASUS\textsubscript{LARGE} checkpoint\footnote{https://huggingface.co/google/pegasus-large} as there is no publicly available PEGASUS\textsubscript{BASE} checkpoint.

\paragraph{GSG*} We use the independent principal strategy of GSG training objective in PEGASUS \cite{zhang2019pegasus} but pre-train \modelname\ with our training corpora. We build this baseline to explore the performance gap between our pre-training objective and GSG.

\section{Results and Analysis}
We focus on low-resource dialogue summarization settings because it is difficult to collect enough training examples. We evaluate \modelname\ with ``All G'', ``All P'', and ``Better ROUGE'' strategies in zero-shot and few-shot settings and compare it to the baselines.

\begin{table*}[htb!]
    \centering
    \footnotesize
    \setlength\tabcolsep{4.5pt}
    \begin{tabular}{l|cccccc|c}
    Model & SAMSum & NYT & Reddit & Stack & Email & TweetSumm & Avg.\\
    \midrule
    \midrule
    T5v1.1     & 9.6/1.6/8.6 &   11.6/1.4/8.7 &  12.3/1.7/9.2 &  15.6/2.4/11.0 & 14.9/2.7/11.1 & 6.0/1.4/5.1 & 11.7/1.9/9.0\\
    PEGASUS   & 27.5/7.6/21.5 & 23.7/3.2/13.2 & 23.1/4.1/13.6 & 26.7/4.8/15.2 & 23.9/5.7/15.3 & 21.8/6.3/16.0 & 24.5/5.3/15.8\\
    GSG*   & 13.3/3.5/12.0 & 17.1/2.4/12.9 & 16.0/2.1/12.5 & 21.2/3.5/15.1 & 21.0/4.2/15.9 &  15.4/2.8/13.1 & 17.3/3.1/13.6\\
    \midrule
    Ours: G & \textbf{41.3}/16.1/30.6 & 21.7/3.7/14.8 & 23.5/4.3/15.7 & 26.3/5.4/16.8 & 26.4/7.1/17.2 & 29.4/8.4/22.1 & 28.1/7.5/19.5\\
    Ours: P & 23.5/7.5/18.6 & 19.8/2.7/12.9 & 20.0/2.9/12.7 & 24.5/4.3/15.0 & 24.3/5.5/15.8 & 22.1/6.7/17.6 & 22.4/4.9/15.4\\
    Ours: BR & \textbf{41.3}/\textbf{16.2}/\textbf{30.9} & \textbf{24.1}/\textbf{4.0}/\textbf{15.4} & \textbf{24.8}/\textbf{4.4}/\textbf{15.9} & \textbf{28.5}/\textbf{5.6}/\textbf{17.6} & \textbf{28.9}/\textbf{7.7}/\textbf{18.0} & \textbf{30.7}/\textbf{10.1}/\textbf{23.4} & \textbf{29.7}/\textbf{8.0}/\textbf{20.2}\\
    \end{tabular}
    \caption{The ROUGE-1/ROUGE-2/ROUGE-L scores of the \modelnamelarge\ with strategy P: ``All P'', G: ``All G'', and BR: ``Better ROUGE'' and compared to T5v1.1\textsubscript{LARGE} and PEGASUS\textsubscript{LARGE} in a zero-shot setting on three datasets: SAMSum, ConvoSumm, and TweetSumm.}
    \label{tab:zero shot res large}
    \vspace{-3mm}
\end{table*}

\subsection{Zero-Shot Results}
In order to evaluate the effectiveness of \modelname, we conduct a zero-shot test on \modelnamelarge\ with all strategies and other baselines. We present the results in Table \ref{tab:zero shot res large}. The ROUGE1-F1, ROUGE2-F1, and ROUGEL-F1 scores are used as the standard evaluation measures for summarization tasks. Our models show impressive performance improvements over the baselines on all downstream datasets. Specifically, \modelnamelarge\ with the ``Better ROUGE'' strategy performs the best overall across all downstream datasets (Average: ROUGE-1/2/L: $29.7/8.0/20.2$), indicating that it benefits from both generated and extractive pseudo summaries and can adapt to various domains. The ``All P'' strategy performs better than the GSG* baseline on most datasets, indicating that our Gap Sentence Selection Plus method can effectively select dialogue turns that provide an accurate dialogue summary. Additionally, the \modelnamelarge\ with ``All G'' and ``Better ROUGE'' strategies demonstrate significant improvement compared to T5v1.1\textsubscript{LARGE} (Average ROUGE2: $+5.6/+6.1$) and PEGASUS\textsubscript{LARGE} (Average ROUGE2: $+2.2/+2.7$), indicating that pre-training with our summary helper is highly beneficial. However, the ``All G'' strategy only performs as well as the ``Better ROUGE'' strategy on the SAMSum dataset (ROUGE-1/2/L/: $41.3/16.1/30.6 \to 41.3/16.2/30.9$), suggesting that the improvement from the summary helper is more pronounced on this particular dataset. This may be due to the similarity between the datasets used to train the helper and the SAMSum dataset, which we discuss further in Sections \ref{sec:helper performance} and \ref{sec:test overlap with traning}. Overall, our models outperform previous methods, such as PEGASUS, in a zero-shot setting, demonstrating their effectiveness and potential for further development.

\subsection{Few-Shot Results}
\label{sec:few shot}
\begin{figure*}[htb]
    \centering
    \includegraphics[trim={0 0 0 0},clip,width=16cm]{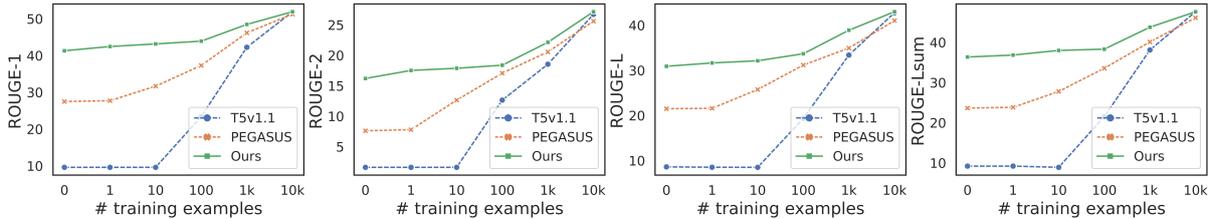}
    \caption{Comparison of T5v1.1\textsubscript{LARGE}, PEGASUS\textsubscript{LARGE}, and \modelnamelarge, fine-tuned with limited training examples on the SAMSum dataset. The training data is within 10,000 examples. The results show that \modelname\ outperforms both PEGASUS and T5v1.1 on all four metrics.}
    \label{fig:exp_few_shot_large}
    \vspace{-3mm}
\end{figure*}
We investigated reducing annotation labor in dialogue summarization tasks by using few-shot dialogue summarization. We report ROUGE1-F1, ROUGE2-F1, ROUGEL-F1, and ROUGELSum-F1 scores to evaluate model performance. Specifically, We fine-tune \modelnamelarge, PEGASUS\textsubscript{LARGE}, and T5v1.1\textsubscript{LARGE} with the first $1/10/100/1K/10K$ training examples from the SAMSum dataset. We show the results of our experiments with varying training data sizes in Figure \ref{fig:exp_few_shot_large}. We found that all models improved with more examples. Among these models, \modelnamelarge\ consistently outperformes both PEGASUS\textsubscript{LARGE} and T5v1.1\textsubscript{LARGE} when trained with a dataset ranging from $0$ to $10,000$ examples. This suggests that our pre-training process helps \modelname\ adapt to downstream tasks more quickly. Additionally, we observed that PEGASUS\textsubscript{LARGE} outperformed T5v1.1\textsubscript{LARGE} due to its pre-training on summarization tasks. Figure \ref{fig:exp_few_shot_large} shows the gap between \modelnamelarge\ and PEGASUS\textsubscript{LARGE} is particularly significant when using fewer than $100$ training examples, indicating better recall capabilities in dialogue summarization for \modelname. Even with only 10 training examples, \modelnamelarge\ achieves higher ROUGE scores than the T5v1.1\textsubscript{LARGE} model trained with 1,000 examples, making it the best option for low-resource dialogue summarization.

\subsection{Effect of Compression Ratio}
\label{sec:compression ratio}
\begin{figure}[htb]
    \centering
    \includegraphics[trim={0 0 0 0},clip,width=7cm]{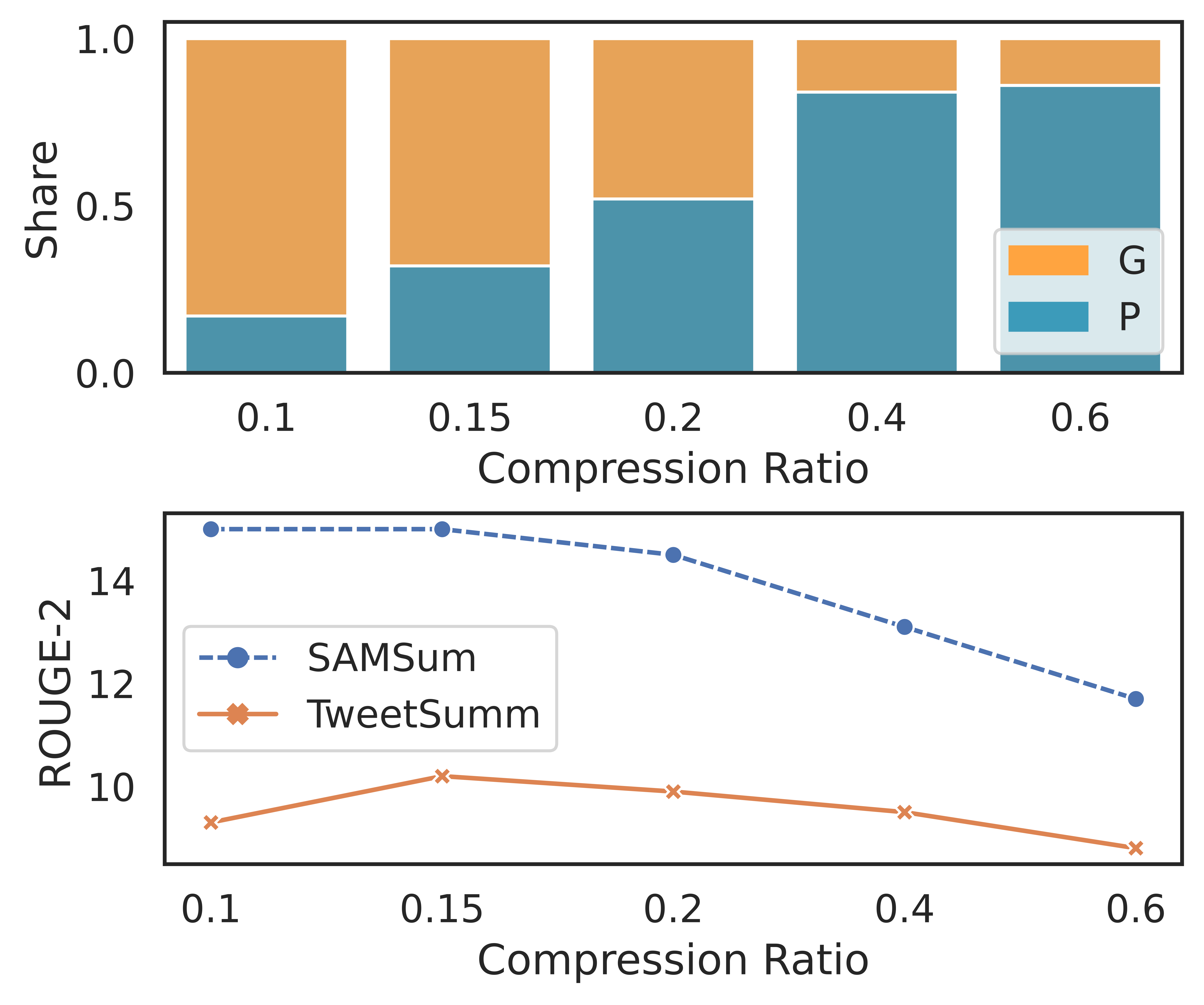}
    \caption{Comparison of compression ratios in \modelnamebase\ using ``Better ROUGE'' strategy. The upper figure reflects the percentage of generated summaries (G) and ``Princial'' (P) at different compression ratios. The performance is measured using the ROUGE2-F1 metric on the SamSum and TweetSumm development sets.}
    \label{fig:exp_compression_rate}
    \vspace{-5mm}
\end{figure}
In GSG+, we can choose a fixed number of turns in the dialogue as a training objective or select turns with a compression ratio. We investigate the compression ratio in a dialogue turn level as the number of selected turns over the number of totals turns in the dialogue ($N_{principal}/N_{dialogue}$). A low compression ratio will select fewer turns in the dialogue as the objective, making pre-training less challenging. However, it tends to have a lower ROUGE1-F1 score with the remaining dialogue turns, meaning the ``Better ROUGE'' strategy selects more generated summaries as the objective. While choosing a high compression ratio will make the pre-training more challenging. Nevertheless, it has a higher ROUGE score compared to generated summaries, leading to more principal under the ``Better ROUGE'' strategy. We show the zero-shot performance on development sets of the SAMSum dataset and TweetSumm dataset with compression rates from $10\%$ to $60\%$ in Figure \ref{fig:exp_compression_rate}. It shows that the model with $15\%$ compression ratio achieves the highest ROUGE-2 score. 

\subsection{Effect of Copying Mechanism}
\label{sec:copying mechanism}
\begin{table}[htb!]
    \centering
    \small
    \begin{tabular}{l|cc}
    ROUGE-1/2/L & All P & w/o copying \\
    \midrule
    \midrule
    SAMSum & \textbf{25.8}/\textbf{8.5}/\textbf{19.7} & 17.7/5.7/15.7 \\
    \midrule
    NYT   & \textbf{21.3}/\textbf{2.7}/\textbf{13.5} & 17.4/2.2/13.4 \\
    Reddit   & \textbf{22.3}/\textbf{3.4}/\textbf{13.8} & 16.3/2.6/13.1 \\
    Stack    & \textbf{25.9}/\textbf{4.5}/\textbf{15.8} & 20.3/3.4/15.1 \\
    Email     & \textbf{26.6}/\textbf{6.1}/\textbf{16.8} & 20.0/3.5/14.7 \\
    \midrule
    TweetSumm     & \textbf{24.1}/\textbf{8.5}/\textbf{19.0} & 19.4/3.8/16.3 \\
    \end{tabular}
    \caption{ROUGE-1/2/L scores of zero-shot setting for \modelnamebase\ with ``All P'' strategy and ``All P'' without copying mechanism on SAMSum, ConvoSumm, and TweetSum.}
    \label{tab:randomly copy res}
    \vspace{-3mm}
\end{table}

\begin{figure}[ht]
    \centering
    \includegraphics[trim={0 0 0 0},clip,width=7cm]{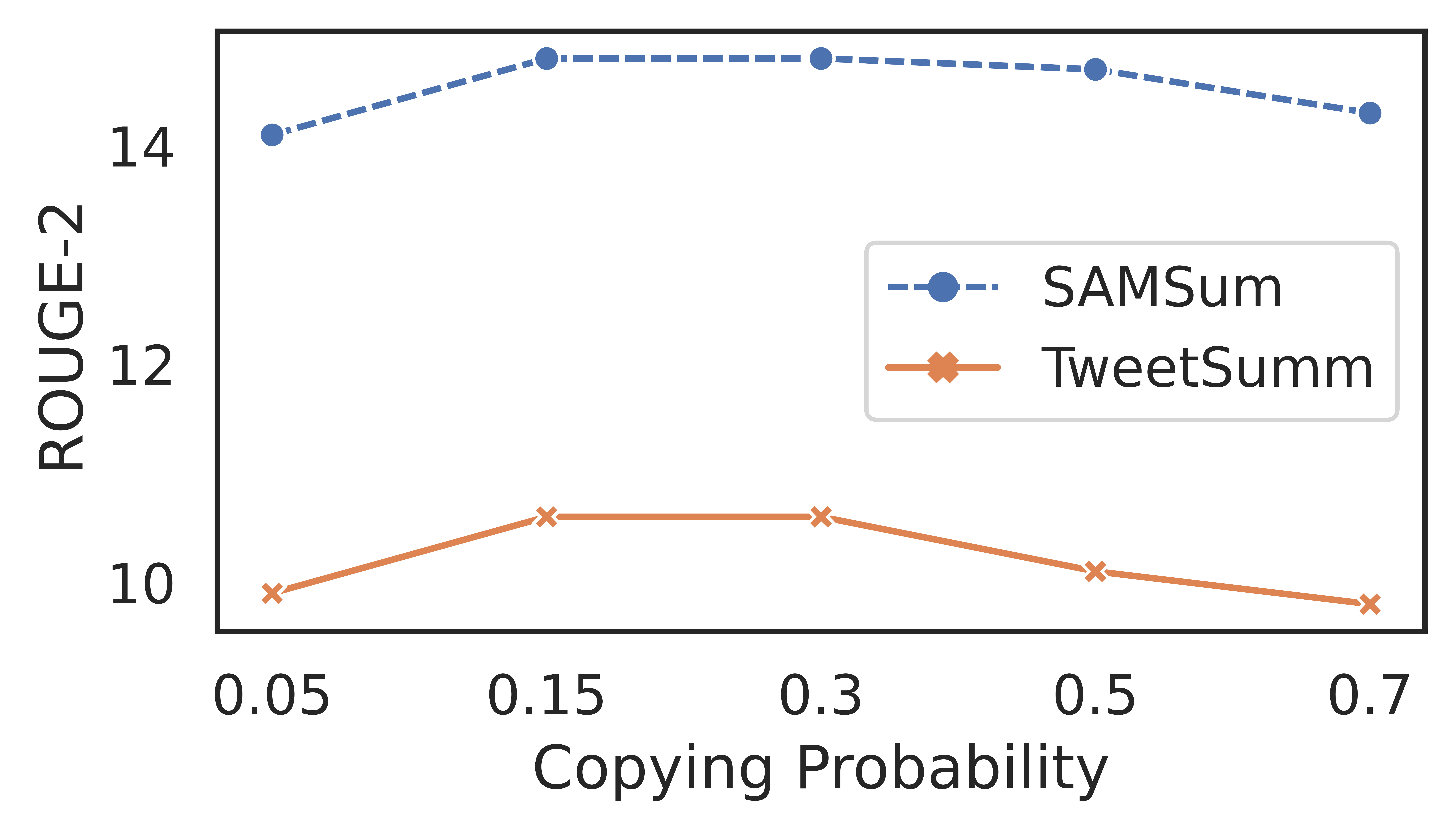}
    \caption{Comparing probabilities of copying selected sentences in the input of the ``Principal'' using the ``Better ROUGE'' strategy. Evaluating performance using the ROUGE2-F1 metric on SamSum and TweetSumm development datasets.}
    \label{fig:exp_copy_prob}
    \vspace{-3mm}
\end{figure}
The copying mechanism is important for dialogues like meetings and medical dialogues because it allows for summarization of entire dialogue through several turns. As shown in Table \ref{tab:randomly copy res}, we compare the performance of the ``All P'' strategy to a scenario where $50\%$ of the selected dialogue turns are retained in the input rather than being removed. In this case, the input for each pre-training example includes the entire dialogue $D$, rather than $D \setminus P$. This leads the model to focus on extractive summarization. We observed that adding a random copy mechanism significantly improved the overall performance. Additionally, we also evaluate the ``Better ROUGE'' strategy with different copying probabilities ranging from $0.15$ to $0.7$. In these experiments, we choose top-2 dialogue turns as principal, which results in $51.9\%$ of pre-training objectives being the principal, and the rest is the generated summary. Figure \ref{fig:exp_copy_prob} shows that leaving $15\%$ of dialogue turns in the principal best enhances the overall quality of dialogue summarization.

\subsection{Comparison Between All G and Summary Helper}
\label{sec:helper performance}
\begin{table}[htb!]
    \centering
    \small
    \begin{tabular}{l|cc}
    ROUGE-1/2/L  & All G & Helper\\
    \midrule
    \midrule
    SAMSum  & \textbf{41.3}/\textbf{16.1}/\textbf{30.6} & 35.8/13.5/27.9\\
    \midrule
    NYT    & \textbf{21.7}/3.7/14.8 & 21.2/4.0/15.2\\
    Reddit    & \textbf{23.5}/\textbf{4.3}/\textbf{15.7} & 20.2/3.5/14.4\\
    Stack     & \textbf{26.3}/\textbf{5.4}/\textbf{16.8} & 25.1/5.0/16.0\\
    Email      & \textbf{26.4}/\textbf{7.1}/\textbf{17.2} & 22.9/5.6/15.2\\
    \midrule
    TweetSumm     & \textbf{29.4}/\textbf{8.4}/\textbf{22.1} & 26.8/6.2/20.8 \\
    \end{tabular}
    \caption{ROUGE-1/2/L scores of zero-shot setting for \modelnamebase\ with ``All G'' strategy and the summary helper on SAMSum, ConvoSumm, and TweetSum.}
    \label{tab:helper performance res}
    \vspace{-3mm}
\end{table}

Since the summary helper model provides the generated summary as an objective candidate and has shown strong capabilities in zero-shot dialogue summarization. As shown in Table \ref{tab:helper performance res}, we compare the helper model to our ``All G'' model in a zero-shot setting. The difference is that we train the ``All G'' model on the pre-training corpora annotated by the helper. We found that the helper model is not on par with our model. While the helper model may have performed well on a particular task (NYT), its overall performance is not as strong as our model. This is because \modelname\ has been extensively trained on various dialogue datasets, which makes it consistently perform well in a wide range of tasks and scenarios.

\subsection{Test-Set Overlap with Pre-Training Corpora}
\label{sec:test overlap with traning}
\begin{table}[htb!]
    \centering
    \small
    \begin{tabular}{l|ccc}
    Threshold & ConvoKit & DS2 & DialogSum \\
    \midrule
    \midrule
    $\geq$ 1.0 & 0\% & 0\% & 0\% \\
    $\geq$ 0.8 & 0\% & 0\% & 0\% \\
    $\geq$ 0.6 & 0\% & 0\% & 1\% \\
    $\geq$ 0.4 & 5\% & 0\% & 3\% \\
    \end{tabular}
    \caption{Percentage of overlap between the SAMSum test set and the datasets used for pre-training. The ConvoKit corpora were comprised of a randomly selected 10\% of the total datafor calculating the similarity.}
    \label{tab:testset overlap res}
    \vspace{-3mm}
\end{table}
In order to ensure a fair comparison, we check for overlap between pre-training and downstream test datasets. This is done by calculating the similarity between all pairs of test set targets in the SAMSum dataset and pre-training documents using the ROUGE2-recall measure, which is calculated as the number of overlapping bigrams divided by the total number of bigrams in the test target. We then count the number of test set examples that have a similarity to any pre-training example above a certain threshold. As shown in Table \ref{tab:testset overlap res}, the overlap between the SAMSum dataset and the datasets used for training the helper and the pre-training datasets is low when the similarity threshold is set between 0.4 and 1.0. This suggests that there is not significant similarity between our test set and the pre-training datasets. It indicates that the improvement in \modelname\ is due to the pre-training process rather than potential test data leakage.

\subsection{Human Evaluation}
\label{sec:human eval}
\begin{table}[htb!]
    \centering
    \small
    \begin{tabular}{l|c}
    & Ratings \\
    \midrule
    \midrule
    T5v1.1\textsubscript{LARGE} & $3.54^{**}$ \\
    PEGASUS\textsubscript{LARGE}    & $3.90^{*}$ \\
    \modelnamelarge\     & $4.04$ \\
    \midrule
    Human-written     & $4.08$ \\
    \end{tabular}
    \caption{Human evaluation results of zero-shot generation. We test the T5v1.1 baseline and the PEGASUS model against \modelname\ with **p < 0.01, *p < 0.05.}
    \label{tab:human eval res}
\end{table}

We evaluate the performance of \modelname\ by conducting human evaluation experiments on Amazon Mechanical Turk. We randomly select 100 examples from the SAMSum dataset to compare summaries generated by our model with those written by humans in the dataset. We choose \modelname\ trained with the ``Better ROUGE'' strategy and generate summaries in a zero-shot setting. Participants are asked to rate the summaries on a scale of 1 to 5, with higher scores indicating better quality. We collect the scores from three participants for each example and report the average scores in Table \ref{tab:human eval res}. A paired t-test is conducted to determine if scores are significantly different between our model and other models. Our results show that \modelname\ could generate summaries of similar quality to human-written summaries without any training data. \modelname\ also gets better ratings than the vanilla T5 and PEGASUS models, which aligns with the results obtained from the automatic evaluation. More information on the human evaluation process can be found in Appendix \ref{sec:human eval details}.

\section{Related Work}
Dialogue summarization is a rapidly growing area of research that focuses on automatically generating concise and informative summaries of conversations \cite{ijcai2022p0764}. Unlike research on traditional documents like news articles \cite{fabbri-etal-2019-multi, ahuja-etal-2022-aspectnews} or scientific papers \cite{lu-etal-2020-multi-xscience, IBRAHIMALTMAMI20221011}, dialogue summarization is particularly relevant in multi-party interactions, such as emails \cite{zhang-etal-2021-emailsum}, meetings \cite{Carletta:83204}, medical dialogues \cite{zeng-etal-2020-meddialog}, and daily chats \cite{chen-etal-2021-dialogsum}. However, many existing methods for dialogue summarization require a large training dataset with annotated summaries. This can be a major barrier to applying these methods in real-world scenarios, particularly in cases with limited or no annotated data available. Our study examines the use of dialogue summarization in low-resource settings to make the process more practical and effortless in various contexts.

Pre-trained Transformer-based \cite{NIPS2017_Vaswani} language models \cite{devlin-etal-2019-bert, Radford2019, NEURIPS2019_Yang} have become increasingly popular in natural language processing tasks for tackling the data shortage problem. However, many of these models have limitations when it comes to dialogue summarization. \citet{zhang2019pegasus} propose PEGASUS, which masks multiple whole sentences and pre-trains sequence-to-sequence models to reconstruct the original text. Built on that, \citet{wan-bansal-2022-factpegasus} improve the sentence selection strategy and add modules for ensuring factuality during fine-tuning to address the problem of factuality in summarization. \citet{Phang2022InvestigatingEE} extend PEGASUS with a modified architecture and long-sequence pre-training to tackle long-input summarization. \citet{he2022z} propose ZCode++, a pre-trained language model optimized for abstractive summarization with improved encoder. However, all these methods rely on the Gap Sentence Selection method, which has limitations for dialogue summarization. In contrast, our approach uses pseudo-summary construction as the pre-training objective, making it possible for zero-shot dialogue summarization.

Another line of work focuses on pre-trained models for dialogues. DialoGPT \cite{zhang-etal-2020-dialogpt} and PLATO \cite{bao-etal-2020-plato}, which are pre-trained on large-scale conversation datasets such as Reddit. For dialogue summarization, \citet{jia-etal-2022-post} post-train pre-trained language models to rephrase dialogues into narratives and then fine-tunes them for summarization. In contrast, our approach follows the T5 model's unified text-to-text format for both pre-training and fine-tuning. \citet{zhong2022DialogLM} train UNILM \cite{Dong2019unilm} with a window-based denoising framework for long dialogue understanding and summarization but do not focus on low-resource settings. \citet{zou-etal-2021-low} propose a pre-training paradigm that pre-trains the encoder and decoder separately in a supervised manner. While our method uses a self-supervised pre-training approach that applies to any dialogue dataset, making it easier to extend to larger pre-training corpora for further improvement.

\section{Conclusion and Future Work}
We present \modelname, a pre-trained encoder-decoder model for zero-shot dialogue summarization in any new domain. We pre-train using a self-supervised approach that generates pseudo-summaries for large dialogue corpora as the pre-training objective. We investigate the impact of various pre-training objective strategies and model sizes on dialogue summarization performance. Our experiments show that \modelname\ outperforms state-of-the-art models on six datasets in a zero-shot setting. Furthermore, \modelname\ can be fine-tuned with only 10 examples to outperform vanilla T5 fine-tuning with 1,000 examples. This makes dialogue summarization more practical and easier to use in various contexts with minimal effort. We plan to extend this method to abstractive summarization tasks to develop a general zero-shot summarization model.

\section{Limitations}
\paragraph{Training Data} Our pre-training data is sourced from 19 existing dialogue datasets. However, it's important to note that these datasets may contain noise, such as harmful content, irrelevant file names, and URL links. Despite utilizing multiple automatic tools to filter out this content during pre-processing, there is still a chance that some noise may be present in our pre-training data. This could potentially impact the performance of \modelname, making it important to monitor and improve the pre-processing steps continuously.

We also know the potential drawbacks of constructing pseudo summaries using the GSG method, which may lead to unnatural summaries for dialogue data. To mitigate this, we introduced the Summary Helper in Section \ref{sec:helper}, which is specifically trained on two dialogue summarization datasets containing natural summaries. This approach enables more realistic pseudo-summaries and enhances zero-shot performance. Although we employ top-m turns as an additional source of pseudo summaries, Figure \ref{fig:exp_compression_rate} illustrates that GSG+ contributes a minor portion of the pseudo summary, with a $0.7$ to $0.3$ ratio between generated and top-m turns. Our method thus minimizes referent and pronoun confusion, ensuring better coherence than solely employing the standard GSG technique.

\paragraph{Training Resource} To improve our model's performance, we employ the ``Better ROUGE'' strategy, which calculates the ROUGE score for both candidates and selects the best one as the final training objective. This data pre-processing process can be pretty time-consuming, taking approximately one day to complete for our pre-training data when utilizing 100 threads. Additionally, we utilize 16 Nvidia V100 GPUs to train our models, which may not be accessible or reproducible for all researchers. This could present a significant obstacle for those looking to replicate or build upon our work.

\paragraph{Test Data} Another potential concern is the test datasets used to evaluate \modelname. The test set size is relatively small, which may not fully represent the breadth of dialogue types that a general dialogue summarization model should be able to handle. This could lead to the model performing well on the test set but not generalizing to other unseen dialogue types. Further, our analysis did not include the assessment of long dialogue summarization, such as lengthy meetings \cite{Carletta:83204, zhong-etal-2021-qmsum, 1198793} or screenplays \cite{chen-etal-2022-summscreen}. However, our study's approach has the potential to handle these scenarios, even though it was not specifically designed for them. By incorporating LongT5 \cite{guo-etal-2022-longt5} or DialogLM \cite{zhong2022DialogLM}, which are known for their ability to process extended input sequences, we expect that they could efficiently tackle this task.

\section{Acknowledgement}
Our gratitude goes out to Microsoft Research for providing us with computational resources. We would also like to thank Kun Qian for valuable discussions and the Columbia NLP and Microsoft Deep Learning Group members for their feedback and discussions. Additionally, we thank the Mechanical Turk workers for conducting the human evaluation.

\bibliography{acl_latex}
\appendix

\section{Implementation Details}
Following \citet{Raffel-t5-2022} and \citet{zhang2019pegasus} to save time and computation, we first conduct ablation experiments on a reduced-size T5v1.1\textsubscript{BASE} model with $250M$ parameters. Then we scale up with the best settings to the final T5v1.1\textsubscript{LARGE} model with $800M$ parameters. We use heuristics to clean up our pre-training corpora. First, we remove dialogues with less than two dialogue turns since they are too short to summarize. Then we remove URLs and emojis in the text. \modelname\ is implemented with Huggingface Pytorch Transformers\footnote{https://github.com/huggingface/transformers is licensed under the Apache License 2.0} \cite{wolf2020transformers}. We split dialogue turns with line breakers in pre-training input and add a ``[Summary]'' prefix. For pseudo summary creation, we use a compression ratio of $0.15$ for the ``Principal.'' This means that for a dialogue with $l$ turns, we select $0.15l$ turns as ``Principal.'' We explore the effect of different compression ratios in Section \ref{sec:compression ratio}. We use Adam \cite{kingma2014adam} with weight decay for pre-training. We truncate dialogue training examples to ensure a maximum length of $512$. Models are pre-trained with batch size $8$ and learning rate $0.00001$ on $16$ Nvidia V100 GPUs until we observe no progress on validation data or up to $5$ epochs. For few-shot experiments in Section \ref{sec:few shot}, we fine-tune models up to $20$ epochs with batch size $8$ and learning rate $0.00005$, and pick the checkpoint with the best validation performance.

\section{Additional Base Model Results}
\begin{table*}[htb!]
    \centering
    \small
    \begin{tabular}{l|cccccc}
    Model & SAMSum & NYT & Reddit & Stack & Email & TweetSumm\\
    \midrule
    \midrule
    T5v1.1\textsubscript{BASE}     & 9.7/1.2/8.6 &   5.8/0.7/4.9 &  8.9/1.2/7.3 &  11.5/1.7/8.9 & 8.4/1.6/7.2 & 6.8/1.0/6.2 \\
    GSG*   & 13.7/4.0/12.6 & 17.9/2.4/13.9 & 15.8/2.2/12.7 & 20.7/3.4/15.5 & 20.8/3.8/15.9 &  17.0/3.2/14.5 \\
    \midrule
    All G    & 39.2/15.2/29.5  & 20.0/3.1/13.7 & 21.4/3.6/14.7 & 24.1/4.9/16.0 & 24.1/6.5/16.0 & 28.3/9.0/22.1 \\
    All P     & 25.8/8.5/19.7 & 21.3/2.7/13.5 & 22.3/3.4/13.8 & 25.9/4.5/15.8 & 26.6/6.1/16.8 & 24.1/8.5/19.0 \\ 
    Better ROUGE     & \textbf{39.6}/\textbf{15.4}/\textbf{30.1} & \textbf{23.1}/\textbf{3.7}/\textbf{15.0} & \textbf{23.1}/\textbf{4.0}/\textbf{15.1} & \textbf{27.3}/\textbf{5.6}/\textbf{17.1} & \textbf{27.0}/\textbf{6.9}/\textbf{17.6} & \textbf{30.3}/\textbf{9.8}/\textbf{23.2} \\
    \end{tabular}
    \caption{The ROUGE-1/ROUGE-2/ROUGE-L scores of the \modelnamebase\ when implemented with different strategies and compared to T5v1.1\textsubscript{BASE} in a zero-shot setting on three datasets: SAMSum, ConvoSumm, and TweetSumm.}
    \label{tab:zero shot res base}
    \vspace{-3mm}
\end{table*}
\begin{figure*}[htb]
    \centering
    \includegraphics[trim={0 0 0 0},clip,width=16cm]{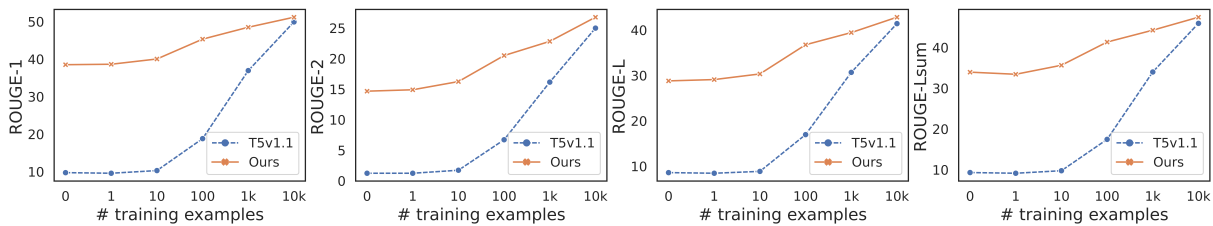}
    \caption{The ROUGE-1, ROUGE-2, ROUGE-L, and ROUGE-LSum scores of low resource dialogue summarization with our best model and T5v1.1. Within 10,000 examples, \modelnamebase\ beats T5v1.1 on all metrics
on SAMSum dataset.}
    \label{fig:exp_few_shot_base}
    \vspace{-3mm}
\end{figure*}
Table \ref{tab:zero shot res base} presents the results of \modelnamebase\ in a zero-shot setting, and Figure \ref{fig:exp_few_shot_base} compares the few-shot results of \modelnamebase\ with those of the T5 base model. These initial results demonstrate the potential for further analysis and optimization of \modelname. Upon comparison with other baselines, it is clear that \modelname\ performs better under both zero-shot and few-shot conditions, outperforming the GSG* model. These results provide valuable insight into the capabilities of \modelname\ and can inform the development of larger models.

\section{Effect of the Dialogue Turns Order in Principal}
\label{sec:effect of order}
\begin{table}[htb!]
    \centering
    \small
    \begin{tabular}{l|cc}
    ROUGE-1/2/L & GSG* (Dialogue) & GSG* (ROUGE) \\
    \midrule
    \midrule
    SAMSum & \textbf{13.7}/\textbf{4.0}/\textbf{12.6} &   13.1/3.7/12.2 \\
    \midrule
    NYT   & \textbf{17.9}/\textbf{2.4}/\textbf{13.9} & 17.6/2.2/13.7 \\
    Reddit   & \textbf{15.8}/2.2/\textbf{12.7} & 15.3/2.2/12.5 \\
    Stack    & \textbf{20.7}/\textbf{3.4}/\textbf{15.5} & 20.1/3.1/15.2 \\
    Email     & \textbf{20.8}/\textbf{3.8}/\textbf{15.9} & 19.8/3.6/15.1 \\
    \midrule
    TweetSumm     & \textbf{17.0}/\textbf{3.2}/\textbf{14.5} & 15.1/2.7/12.8 \\
    \end{tabular}
    \caption{ROUGE-1/2/L scores of zero-shot setting for \modelnamebase\ with GSG* and unordered GSG* on SAMSum, ConvoSumm, and TweetSum.}
    \label{tab:order of sentence in principal res}
\end{table}
We could use two possible orders to align the dialogue turns in the principal. The first order is to align the text with the ROUGE1-F1 score. The second order is to align the principal with the order in the original dialogue. This means that the principal will be arranged in the same order as in the original dialogue, without rearrangement. This option helps preserve the original flow and structure of the dialogue. We compare these two orders of principal in the GSG* baseline. As shown in Table \ref{tab:order of sentence in principal res}, the results suggest that keeping the order in the original dialogue helps improve zero-shot performance as it provides a more nuanced understanding of the dialogue. We choose this order for all our models.

\section{Pre-training Steps}
\begin{figure*}[htb]
    \centering
    \includegraphics[trim={0 0 0 0},clip,width=16cm]{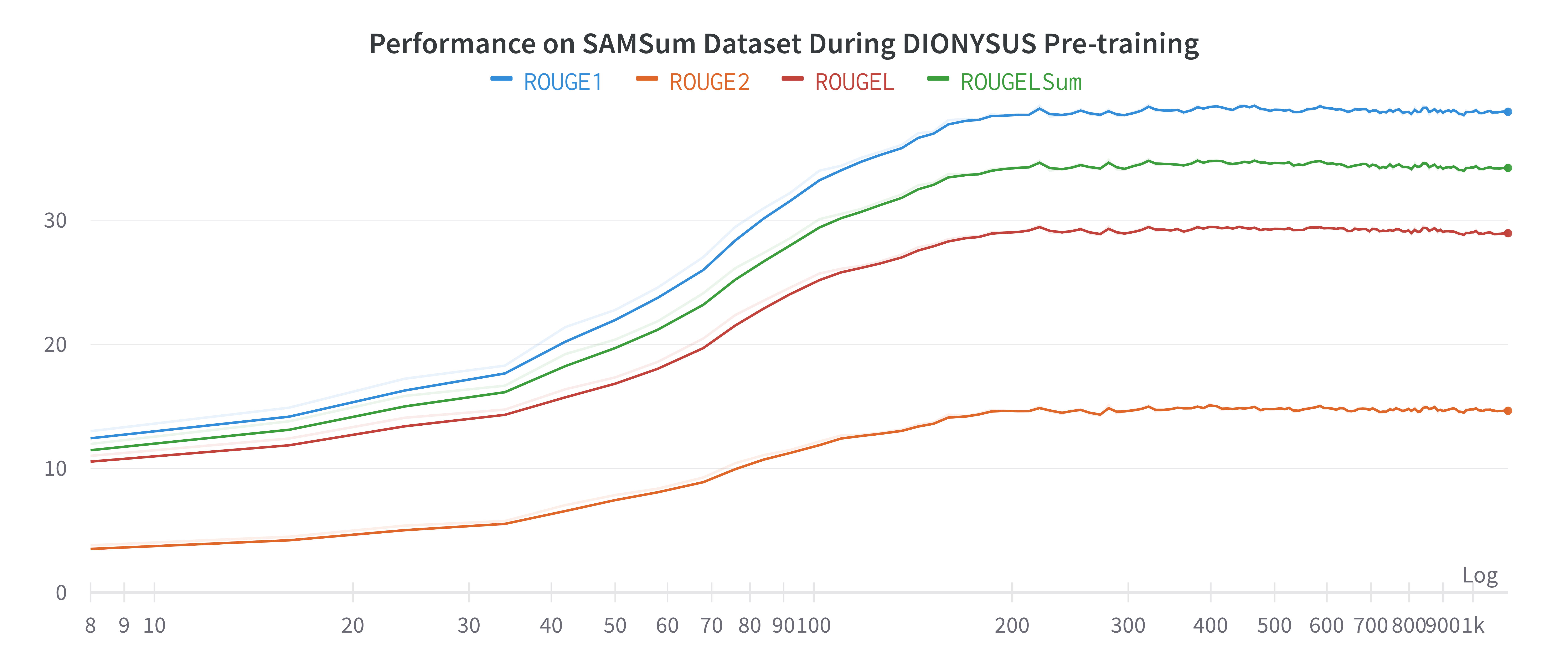}
    \caption{Performance of \modelname\ on the SAMSum dataset during pre-training process.}
    \label{fig:pre_training_steps}
\end{figure*}
To evaluate the performance of \modelname\ during pre-training, we measured the ROUGE1-F1, ROUGE2-F1, ROUGEL-F1, and ROUGELSum-F1 scores on the SAMSum dataset in Figure \ref{fig:pre_training_steps}. We keep track of the model's progress by logging its performance every 1,000 training steps. This allows us to monitor the model's improvements over time and confirm that it is learning effectively.

\section{Example Model Outputs}
In order to evaluate the performance of \modelname, we randomly selected model output examples from both the SAMSum dataset and the TweetSumm dataset. We report these examples with their corresponding gold summaries in Tables \ref{tab:example_samsum} and \ref{tab:example_tweetsumm}. The gold summaries served as a benchmark for our model's output, allowing us to compare and estimate the quality of the generated summaries. We found that \modelname\ could generate zero-shot summaries on par with those written by humans. However, we also identified factual errors in the generated summaries, such as misunderstandings of the subject matter. These errors suggest room for improvement in \modelname, and we plan to address this issue in future work.
\begin{table*}[htb!]
    \centering
    \small
    \begin{tabular}{l|l}
    \toprule
    Example &SAMSum \\
    \midrule
    \multirow{9}{*}{Dialogue\#1} &\cellcolor{mygray}Dzuka: Until further notice, the staff meeting will be held at 8:30 instead of 8:00.\\
    &\cellcolor{mygray}Please change the calendar for everyone. Thanks.\\
                                &Anna: No problem. Why the change\\
                                &\cellcolor{mygray}Dzuka: We had a few that never make it on time. I'm hoping this will encourage more participation.\\
                                &Anna: Could be just the opposite!\\
                                &\cellcolor{mygray}Dzuka: We'll give it a try.\\
                                &Anna: Sure, no problem.\\
                                &\cellcolor{mygray}Dzuka: I'll let you know if it changes again. Thanks.\\
                                &Anna: NP\\
    \midrule
    \multirow{2}{*}{Gold} &The stuff meeting is postponed from 8.00 to 8.30 to encourage more participation.\\
    &Dzuka will inform Anna if it changes again.\\
    \midrule
    \multirow{2}{*}{\modelname} &The staff meeting will be held at 8:30 instead of 8:00.\\
    &Dzuka hopes this will encourage more participation and will let Anna know if it changes again.\\
    \midrule
    \midrule
    \multirow{10}{*}{Dialogue\#2} &\cellcolor{mygray}Jane: Hello\\
                                &Vegano Resto: Hello, how may I help you today?\\
                                &\cellcolor{mygray}Jane: I would like to make a reservation for 6 people, tonight around 20:00\\
                                &Vegano Resto: Let me just check. Ah, I'm afraid that there is no room at 20:00.\\ &However, I could offer you a table for six at 18:30 or at 21:00. Would either of those times suit you?\\
                                &\cellcolor{mygray}Jane: Oh dear. Let me just ask my friends.\\
                                &Vegano Resto: No problem.\\
                                &\cellcolor{mygray}Jane: 21:00 will be ok.\\
                                &Vegano Resto: Perfect. So tonight at 21:00 for six people under your name.\\
                                &\cellcolor{mygray}Jane: great, thank you!\\ 
    \midrule
    Gold &Jane made a 9 PM reservation for 6 people tonight at Vegano Resto.\\
    \midrule
    \multirow{2}{*}{\modelname} &The restaurant has no room for six people at 20:00 and offers Jane a table for six at 18:30 or 21:00.\\
    &Jane asks her friends to make a reservation at 21:00.\\
    \midrule
    \midrule
    \multirow{6}{*}{Dialogue\#3} &\cellcolor{mygray}Mia: Hi Dad! I need a hand with repairing the bathroom door.\\
                                &William: Hi! What happened?\\
                                &\cellcolor{mygray}Mia: Nothing. I can't open/close it properly. It's sort of sagging.\\
                                &William: I see. I'll drop by after work and take a look.\\
                                &\cellcolor{mygray}Mia: Thank you so much! Love you!\\
                                &William: I love you too.\\
    \midrule
    Gold &Mia's dad William will come to her place after work to repair her bathroom door.\\
    \midrule
    \modelname\ &The bathroom door is sagging. William will drop by after work and take a look.\\
    \bottomrule
    \end{tabular}
    \caption{Example zero-shot outputs on the SAMSum dataset.}
    \label{tab:example_samsum}
\end{table*}
\begin{table*}[htb!]
    \centering
    \small
    \begin{tabular}{l|l}
    \toprule
    Example &\textsc{TweetSumm} \\
    \midrule
    \multirow{17}{*}{Dialogue\#1} &\cellcolor{mygray}@549761: My BITS service resets it's Startup type from disabled to automatic.\\
    &\cellcolor{mygray}It leeches on to my bandwidth like crazy. Please provide a solution.\\
    &@MicrosoftHelps: Hi. We hear you. We'd like to check what happened prior to this issue?\\
    &What Windows version are you using? Let us know.\\
    &\cellcolor{mygray}@549761: I am using Windows 10 Home Single Language. Nothing specific happened prior to this issue.\\
    &\cellcolor{mygray}Just the service used to leech on to bandwidth (it claims to use idle network but doesn't).\\
    &\cellcolor{mygray}I want it to stop from resetting Startup type from disabled to automatic.\\
    &@MicrosoftHelps: Thanks for the info. For us to isolate your concern,\\
    &let's try the troubleshooting steps 1/2\\
    &https://t.co/3qcAsLFkaY listed in this link:\\
    &https://t.co/IBZ1MaTm1l. Kindly check the post of Jesinta Rozario.\\
    &@MicrosoftHelps: Hi, Pratik. How's it going?\\
    &Please let us know if you need further assistance. We're here for you.\\
    &\cellcolor{mygray}@549761: Hi. The service still becomes running after disabling(after a few days).\\
    &\cellcolor{mygray}What can be the reason for the service switching it's startup type?\\
    &@MicrosoftHelps: In that case, we suggest contacting Answer Desk: https://t.co/9Ouw33YVZI\\
    &to further assist you with your concern. Let us know how it goes.\\
    &@MicrosoftHelps: Hello, Pratik! Were we able to resolve your concern?\\
    &If no, we're just one tweet away if you have other concerns.\\
    &If yes, please send us your feedback about your experience with our support here: https://t.co/CczzJgTng1.\\
    \midrule
    \multirow{3}{*}{Gold} &Customer is complaining about the BITS service for resetting startup type from disabled mode to automatic.\\
    &Agent suggests to try out some troubleshooting steps by following the shared URL\\
    &and reach out Answer desk team for further assistance.\\
    \midrule
    \multirow{3}{*}{\modelname} &The BITS service leeches on to the bandwidth like crazy. \\
    &Pratik wants it to stop from resetting Startup type from disabled to automatic.\\
    &MicrosoftHelps suggests checking the post of Jesinta Rozario.\\
    \midrule
    \midrule
    \multirow{20}{*}{Dialogue\#2} &\cellcolor{mygray}@471404: Please bring security back to the Hall Green store.\\
    &\cellcolor{mygray}@471404: The store is getting a more an more uncomfortable vibe, not alone on this either!\\
    &@Tesco: Hi there, sorry to be a pain but can you confirm which Hall Green store this is? TY - Reece \\
    &\cellcolor{mygray}@471404: It's the Hall Green store right next to the train station.\\
    &\cellcolor{mygray}Hoping you haven't removed security from the others too now...\\
    &@Tesco: Hi, can you please confirm what you mean by "uncomfortable vibe"? - Nick\\
    &\cellcolor{mygray}@471404: Well there's pretty obvious shop lifters regularly, \\
    &\cellcolor{mygray}and today we had a man clearly intoxicated screaming and randomly asking people things. \\
    &@Tesco: Yes the express store! Thanks aswell. I'd review the CCTV from when security were removed.\\
    &If customers can see the changes you will too!\\
    &@Tesco: Hi there. I have spoken to the store.\\
    &They have had a few problems recently and are looking into improving security. Thanks - Ian\\
    &\cellcolor{mygray}@471404: Thank you again. I often worry for the staff as it is becoming a hot spot for undesirables.\\
    &\cellcolor{mygray}The homeless aren't the issue to save confusion!\\
    &@Tesco: Hi there, thank you for bringing this to our attention\\
    &the last thing we want is our customers to feel unsafe.\\
    &Thank you - Brooke\\
    &\cellcolor{mygray}@471404: No thank you for taking it seriously  here's hoping the store gets back to normal soon!\\
    &@Tesco: Hi there, I'm glad one of my colleagues has dealt with the issue.\\
    &Enjoy the rest of your weekend - Rian\\
    \midrule
    Gold &The customer is complaining that he facing some uncomfortable vibe.\\
    &The agent confronted the customer saying that they had a few problems recently\\
    &and they are looking into improving security.\\
    \midrule
    \multirow{4}{*}{\modelname} &The store is getting a more an more uncomfortable vibe.\\
    &Nick asks Tesco to bring security back to the Hall Green store and confirms the location.\\
    &Nick also tells Tesco the Express store has had some problems recently\\
    &and is looking into improving security.\\
    \bottomrule
    \end{tabular}
    \caption{Example zero-shot outputs on the \textsc{TweetSumm} dataset.}
    \label{tab:example_tweetsumm}
\end{table*}

\section{Human Evaluation Details}
\label{sec:human eval details}
\begin{figure*}[htb]
    \centering
    \includegraphics[trim={0 0 0 0},clip,width=16cm]{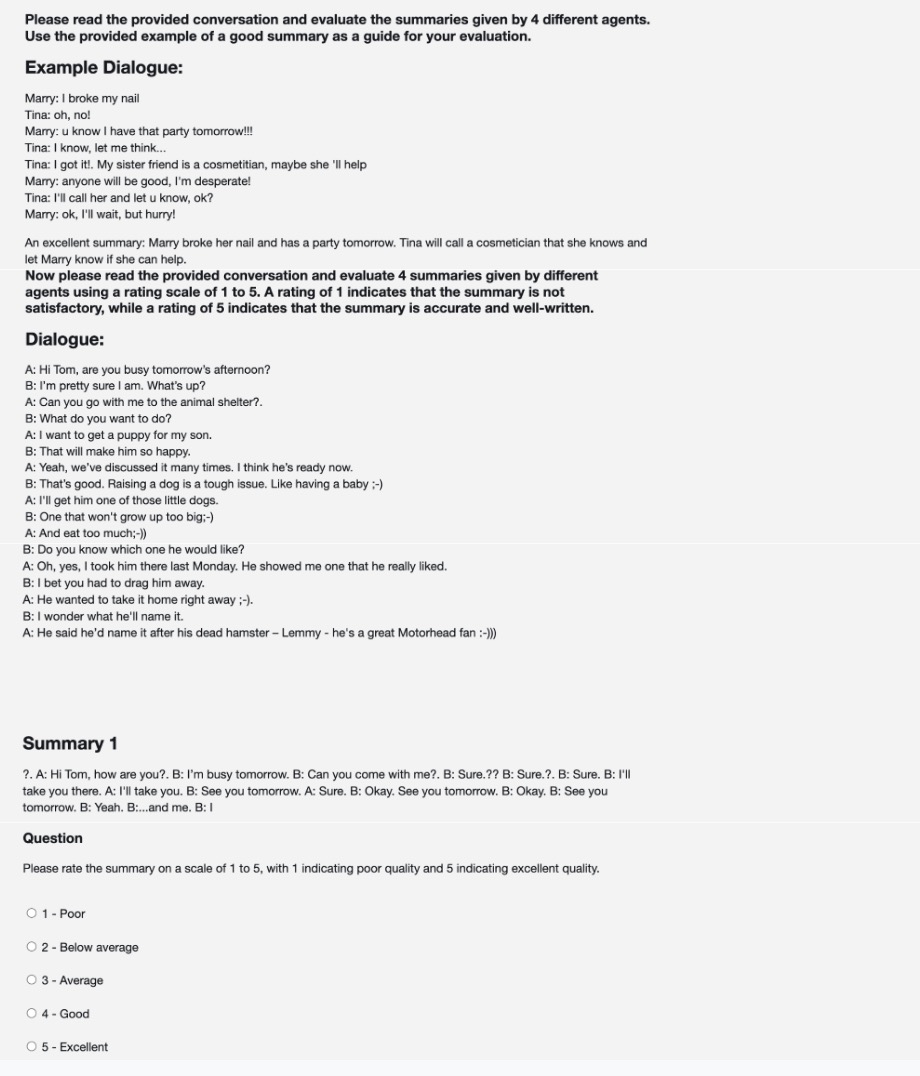}
    \caption{A screenshot of the human evaluation on Amazon Mechanical Turk.}
    \label{fig:human_eval_screenshot}
\end{figure*}
In our human evaluation experiments, we utilized the task template shown in Figure \ref{fig:human_eval_screenshot}. Mechanical workers were instructed to rate four summaries for a given dialogue on a scale of 1 (poor) to 5 (excellent). To minimize bias, we provided a dialogue with its corresponding gold summary as an example of a high-quality summary. The summaries were presented in a randomized order for each task to prevent order bias. Three different workers independently completed each task, and the median score across all workers was retained for each summary. Participants were compensated with 0.3 USD per task, and we implemented the following qualifications for worker selection to ensure a high level of quality: (1) HIT approval rate for all requesters' HITs is greater than 90\%. (2) Location is one of AU, NZ, GB, and US. (3) Number of HITs approved is greater than 100.


\end{document}